\documentclass[conference]{IEEEtran}
\IEEEoverridecommandlockouts
\usepackage{cite}
\usepackage{amsmath,amssymb,amsfonts}
\usepackage{algorithmic}
\usepackage{graphicx}
\usepackage{textcomp}
\usepackage{xcolor}
\usepackage{algorithm}
\usepackage{array}
\usepackage{multirow}

\DeclareMathOperator*{\argmin}{argmin}

\newcommand{\methodname}{{\tt{MultiBOS-AFL}}}
\newcommand{\interBPA}{{\tt{InterBPA}}}
\newcommand{\intraBA}{{\tt{IntraBA}}}
\def\BibTeX{{\rm B\kern-.05em{\sc i\kern-.025em b}\kern-.08em
    T\kern-.1667em\lower.7ex\hbox{E}\kern-.125emX}}
\begin{document}

\title{Multi-Session Budget Optimization for Forward Auction-based Federated Learning}

\author{Xiaoli Tang and Han Yu}

\maketitle

\begin{abstract}
Auction-based Federated Learning (AFL) has emerged as an important research field in recent years. The prevailing strategies for FL model users (MUs) assume that the entire team of the required data owners (DOs) for an FL task must be assembled before training can commence. In practice, an MU can trigger the FL training process multiple times. DOs can thus be gradually recruited over multiple FL model training sessions. Existing bidding strategies for AFL MUs are not designed to handle such scenarios. 
Therefore, the problem of multi-session AFL remains open. 
To address this problem, we propose the \underline{Multi}-session \underline{B}udget \underline{O}ptimization \underline{S}trategy for forward \underline{A}uction-based \underline{F}ederated \underline{L}earning (\methodname{}). Based on hierarchical reinforcement learning, \methodname{} jointly optimizes inter-session budget pacing and intra-session bidding for AFL MUs, with the objective of maximizing the total utility.
Extensive experiments on six benchmark datasets show that it significantly outperforms seven state-of-the-art approaches. On average, \methodname{} achieves 12.28\% higher utility, 14.52\% more data acquired through auctions for a given budget, and 1.23\% higher test accuracy achieved by the resulting FL model compared to the best baseline.
To the best of our knowledge, it is the first budget optimization decision support method with budget pacing capability designed for MUs in multi-session forward auction-based federated learning.
\end{abstract}


\section{Introduction}
\label{sec:introduction}
\begin{figure*}[t!]
\centering
\includegraphics[width=1\linewidth]{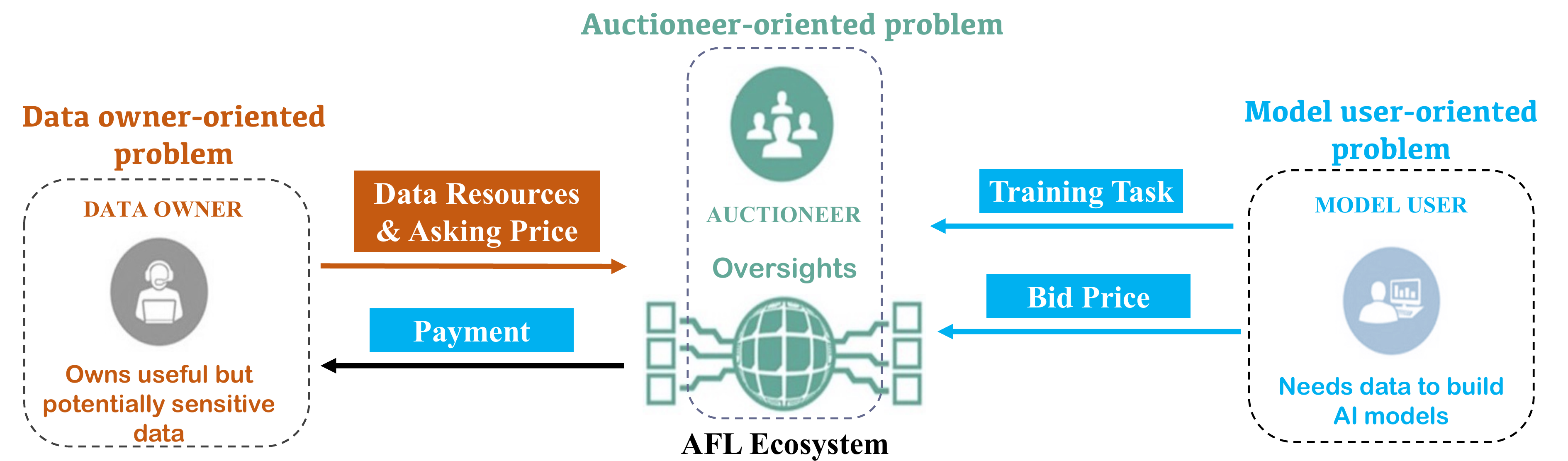}
\caption{An overview of auction-based federated learning (AFL).}
\label{fig:afl}
\end{figure*}
Federated Learning (FL) \cite{yang2019federated,Yang-et-al:2020FL,Goebel-et-al:2023} has emerged as a useful collaborative machine learning (ML) paradigm. In contrast to the traditional ML paradigm, FL enables collaborative model training without the need to expose local data, thereby enhancing data privacy and user confidentiality. Prevailing FL methods often assume that data owners (DOs, a.k.a, FL clients) are ready to join FL tasks by helping model users (MUs, a.k.a, FL servers) train models. In practice, this assumption might not always hold due to DOs' self-interest and trade-off considerations. To deal with this issue, the domain of auction-based federated learning (AFL) has emerged \cite{jiao2019auction,deng2021fair,zhang2021incentive}.

As shown in Figure \ref{fig:afl}, in the context of AFL, the main actors include the auctioneer, DOs and MUs. The auctioneer functions as an intermediary, facilitating the flow of asking prices from DOs and MUs. MUs then determine their bid prices to be submitted to the auctioneer.
The auctioneer then consolidates the auction outcomes and informs the DOs and MUs about the match-making results. The auctioneer undertakes a pivotal role in orchestrating the entire auction process, managing information dissemination, and ultimately determining the auction winners. Once FL teams have been established through auctions, they can carry out collaborative model training following standard FL protocols.

AFL methods can be divided into three categories \cite{tang2023utility}: 1) data owner-oriented (DO-oriented), 2) auctioneer-oriented, and 3) model user-oriented (MU-oriented). DO-oriented AFL methods focus on helping DOs determine the amount of resources to commit to FL tasks, and set their respective reserve prices for profit maximization. Auctioneer-oriented AFL methods investigate how to optimally match DOs with MUs as well as provide the necessary governance oversight to ensure desirable operational objectives can be achieved (e.g., fairness, social cost minimization). MU-oriented AFL methods examine how to help MUs select which DOs to bid and for how much, in order to optimize key performance indicators (KPIs) within budget constraints, possibly in competition with other MUs.

This paper focuses on MU-oriented AFL. The prevailing approach in this domain requires that the budget of an MU shall be maximally spent to recruit the entire team of necessary DOs before FL model training can commence \cite{jiao2020toward, zeng2020fmore, ying2020double, le2020auction, le2021incentive, zhang2021incentive, roy2021distributed, deng2021fair, zhang2022auction, zhang2022online,tang2023utility}.
In practice, throughout the FL model training process, an MU can recruit DOs over multiple training sessions. This is especially useful in continual FL \cite{Yoon-et-al:2021} settings where DOs' local data are continuously updated over time.
Existing AFL approaches are generally designed to optimize KPIs within a single auctioning session. For instance, Fed-Bidder \cite{tang2023utility} takes into account MUs' limited budgets, the suitability of DOs, prior auction-related knowledge (e.g., the data distribution of the DOs, the probability of the MU winning an auction) to design optimal bidding functions. MARL-AFL \cite{tang2023competitive} adopts a multi-agent system approach to steer MUs who bid strategically towards an equilibrium with desirable overall system characteristics. In this method, each MU is represented by its agent. 
Existing MU-oriented AFL methods cannot be directly applied in multi-session AFL scenarios, especially in scenarios with multiple MUs competing to bid for DOs from a common pool of candidates. This is primarily due to the limitation that they are unable to perform budget pacing, which pertains to the strategic dispersion of a limited overall budget across multiple AFL sessions to achieve optimal KPIs over a given time frame. 

To bridge this important gap, we propose a first-of-its-kind \underline{Multi}-session \underline{B}udget \underline{O}ptimization \underline{S}trategy for forward \underline{A}uction-based \underline{F}ederated \underline{L}earning (\methodname{}). It is designed to empower an MU with the ability to dynamically allocate its limited budget over multiple AFL DO recruitment sessions, and then optimize the distribution of budget for each session among DOs through effective bidding. The ultimate goal is to maximize the MU's winning utility. \methodname{} is grounded in Hierarchical Reinforcement Learning (HRL) \cite{HRL2021} to effectively deal with the intricate decision landscape and the absence of readily available analytical remedies. Specifically, \methodname{} consists of two agents for each MU: 1) the Inter-Session Budget Pacing Agent (\interBPA{}), and 2) the Intra-Session Bidding Agent (\intraBA{}). 
For each auctioning session, each MU's \interBPA{} opportunistically determines how much of the total budget shall be spent in this session based on jointly considering the quantity and quality of the currently available candidate DOs, as well as bidding outcomes from previous sessions. Then, the MU's \intraBA{} determines the bid price for each data resource offered by DOs in the AFL market within the current session budget. 

To the best of our knowledge, \methodname{} is the first budget optimization decision support method with budget pacing capability designed for MUs in multi-session forward auction-based federated learning.
Extensive experiments on six benchmark datasets show that it significantly outperforms seven state-of-the-art approaches. On average, \methodname{} achieves 12.28\% higher utility, 14.52\% more data acquired through auctions for a given budget, and 1.23\% higher test accuracy achieved by the resulting FL model compared to the best baseline.

\section{Related Work}
\label{sec:related_work}
Existing AFL approaches can be divided into two groups: 1) methods for the entire AFL ecosystem, and 2) methods for a single AFL MU.

\subsection{Methods for the Entire AFL Ecosystem}
Methods under this category are designed to achieve the overall objectives of an AFL ecosystem for all participating MUs (e.g., social welfare maximization, social cost minimization, total utility maximization). They often draw inspiration from double auctions or combinatorial auctions to strategically determine the optimal matching between MUs and DOs, along with associated pricing.

\subsubsection{Double Auction-based Methods}
The techniques rooted in double auction \cite{li2019credit, krishnaraj2022future, zavodovski2019decloud, hong2020optimizing} come into play when there are multiple DOs and MUs involved in AFL. Here, DOs offer their data resources, while MUs respond with their respective bids. The auctioneer then orchestrates the process to determine the optimal auction outcomes.

\subsubsection{Combinatorial Auction-based Methods}
These approaches \cite{bahreini2018envy, gao2019auction, jiao2018social, jiao2019auction, yang2020task} prove effective when data resources are bundled as combinations, inviting competitive bids from MUs seeking those specific combinations.
For instance, Yang et al. \cite{yang2020task} proposed a multi-round sequential combination auction model catering to the heterogeneous resource requirements of model users and limited data resources of data owners. They establish a dynamic process where bids are submitted in rounds, ultimately leading to optimized winners and pricing. In a different vein, Krishnaraj et al. \cite{krishnaraj2022future} introduced an iterative double auction approach tailored for trading computing resources. Here, iterative optimization tasks, aligned with pricing rules, shape the allocation of winners and pricing through multiple iterations.

\subsection{Methods for a Single AFL Model User}
This second category can be further divided into two subcategories: i) reverse auction-based methods, and ii) forward auction-based methods.

\subsubsection{Reverse Auction-based Methods} 
Developed primarily for monopoly AFL markets where there is only one MU facing multiple DOs, reverse auction-based methods \cite{jiao2020toward, zeng2020fmore, ying2020double, le2020auction, le2021incentive, zhang2021incentive, roy2021distributed, deng2021fair, zhang2022auction, zhang2022online} address the challenge of DO selection through reverse auctions. The key idea of these methods is to optimally resolve the DO selection problem, targeting the maximization of KPIs specific to the target MU. Particularly relevant in scenarios where disparate DOs vie for the attention of a sole MU, these methods have progressed by integrating diverse mechanisms such as graph neural networks, blockchains, and reputation assessment. A notable example is the RRAFL approach \cite{zhang2021incentive}, where blockchain and reputation mechanisms intertwine with reverse auction. In this scenario, the MU initiates a training task, triggering DOs to bid. Winning DOs are chosen based on their reputation reflecting their reliability and quality, gauged through historical records stored on the blockchain for added data integrity assurance.

\subsubsection{Forward Auction-based Methods} 
Forward auction-based methods are designed for situations where multiple MUs compete for the same pool of DOs \cite{tang2023utility}. The key idea of these methods lies in determining the optimal bidding strategy for MUs. The goal is to maximize model-specific key performance indicators. A notable example is Fed-Bidder \cite{tang2023utility} which assists MUs to determine their bids for DOs. It leverages a wealth of auction-related insights, encompassing aspects like DOs' data distributions and suitability to the task, MUs' success probabilities in ongoing auctions and budget constraints. However, this method ignores the complex relationships among MUs, which are both competitive and cooperative. To deal with this issue, Tang et al. \cite{tang2023competitive} model the AFL ecosystem as a multi-agent system to steer MUs to bid strategically toward an equilibrium with desirable overall system characteristics.

\methodname{} falls into the forward auction-based methods category. Distinct from existing methods which focus on optimizing the objectives within a single auctioning session, it is designed to solve the problem of multi-session AFL budget optimization.

\section{Preliminaries}
\label{sec:preliminary}
\subsection{Multi-Session Budget Constrained AFL Bidding}
During the course of FL model training, an MU can initiate the FL training procedure (i.e., a \textit{training session}) on multiple occasions, with the aim of recruiting DOs to improve model performance. Consider the scenario of multiple banks engaging in FL. The dynamic nature of user data within these banks sets in motion a perpetual cycle of updates, with continually refreshed data stored locally by each bank. As a result, these banks systematically engage in repeated sessions of federated model training periodically, during which the standard FL training protocol is followed. 

Let $S$ denote the number of training sessions for the target MU, who has a budget $B$ for all training sessions $[S]$.
In each FL training session $s$ ($s \in [S]$), there are $C_s$ available qualified DOs, which can help train the FL model of the target MU. Each DO $i \in [C_s]$ possesses a private dataset $D_i=\{(\boldsymbol{x}_j, y_j)\}_{j=1}^{|D_i|}$. Following the approach proposed in \cite{zhang2022online}, we assume that the data of each qualified DO $i$ become gradually available over time. Each new data resource from a DO can trigger the following auction process:
\begin{enumerate}
    \item \textbf{Bid Request}: When a qualified DO $i \in [C_s]$ becomes available to join FL training, an auction is initiated. A bid request containing information about the DO (e.g., identity, data quantity, etc.) and the reserve price (i.e., the lowest price the DO is willing to accept for selling the corresponding resources) is generated and sent to the auctioneer \cite{vincent1989bidding}.
    \item \textbf{Bid Request Dissemination}: The auctioneer disseminates the received bid request to MUs who are currently seeking to recruit DOs.
    \item \textbf{Bidding Decision}: Each MU evaluates the potential value and cost of the received bid request and decides whether to submit a bid price or not, based on its bidding strategy.
    \item \textbf{Bid Response}: If an MU decides to bid, it calculates a bid price for the given DO and submits it to the auctioneer.
    \item \textbf{Outcome Determination}: Upon receiving bids from MUs, the auctioneer computes the market price based on the given auction mechanism (e.g., the generalised second-price auction mechanism). It then compares the market price with the reserve price set by each DO. If the market price is lower than the reserve price, the auctioneer terminates the auction and informs the DO to initiate another auction for the same resources. Otherwise, the auctioneer informs the winning MU about the cost (i.e., the market price) $p_s^i$ it needs to pay, and informs the DO about the winning MU it shall join.
\end{enumerate}

The auction process described above can be conducted by the auctioneer for each bundle of eligible data resources owned by each DO $i \in [C_s]$. 
Each DO may possess multiple types of data resources suitable for various FL model training tasks. It can also commit only a portion of its data resources to train a specific FL model. When auctioning process for session $s$ has completed or the MU has exhausted its budget, it initiates FL model training with the recruited DOs. Each MU pays the corresponding market prices to the DOs it has recruited.

\subsection{Federated Learning with Recruited Data Owners}
After the auction-based DO recruitment process, the MU triggers the FL training process with the recruited DOs in session $s$. Specifically, the FL process operates through communication between the recruited DOs and the target MU in a round-by-round manner. In each training round $t$ in session $s$, the target MU broadcasts the current global model parameters $\boldsymbol{w}^{t-1}_{s}$ to the recruited DOs. Upon receiving $\boldsymbol{w}^{t-1}_{s}$, each DO $i$ performs a local update to obtain $\boldsymbol{w}^{t}_{s,i}$ based on its private data $D_i$, guided by the following objective function:
\begin{equation}
\label{eq:fl_slient_loss}
\argmin_{\boldsymbol{w}_{s,i}^{t}} \mathbb{E}_{(\boldsymbol{x}, y) \sim D_i} [\mathcal{L}(\boldsymbol{w}^{t}_{s,i};( \boldsymbol{x}, y)].
\end{equation}
$\mathcal{L}(\cdot)$ represents the loss function, which depends on the FL model aggregation algorithm and the current global model parameters $\boldsymbol{w}^{t-1}_s$. For instance, FedAvg \cite{mcmahan2017communication} calculates $\boldsymbol{w}^{t}_{s,i}$ by employing SGD \cite{robbins1951stochastic} for a certain number of epochs using the cross-entropy loss. At the end of round $t$, $i$ sends its optimized parameters $\boldsymbol{w}^{t}_{s,i}$ to the target MU. The global model is then updated by aggregating these parameter updates from the DOs: 
\begin{equation}
\label{eq:fl_global_parameter_update}
\boldsymbol{w}^{t}_s = \sum_{i} \frac{|D_i|}{\sum_{i}|D_i|} \boldsymbol{w}^{t}_{s,i}.
\end{equation}
$\sum_{i}|D_i|$ denotes the total number of data samples of all the recruited DOs in session $s$.

Let $v_s^i$ denote the reputation of DO $i \in [C_s]$ \cite{shi2023fairfed} and $x_s^i \in \{0,1\}$ denote whether the target MU wins $i$. Then, the goal of the target MU across $S$ sessions is to maximize the total utility of winning DOs\footnote{Following \cite{tang2023competitive,tang2023utility,zhang2021incentive,zhang2022online,zhang2022auction,zhan2020experience}, maximizing the total utility is equivalent to optimizing the performance of the global FL model obtained by the target MU.} under a budget $B$, which can be formulated as:
\begin{equation}
\label{eq:goal}
\begin{aligned}
& \max \sum_{s \in [S]} \sum_{i \in [C_s]} x_s^i \times v_s^i, \\
& s.t. \quad \sum_{s \in [S]} \sum_{i \in [C_s]} x_s^i \times p_s^i \leq B, 
\end{aligned}
\end{equation}
Following \cite{shi2023fairfed}, we calculate the reputation of each DO based on the Shapley Value (SV) \cite{shapley1953value} technique and Beta Reputation System  (BRS) \cite{josang2002beta}. 

We start by adopting the SV approach to calculate the contribution $\phi_i$ of each DO $i$ during each training round towards the performance of the resulting FL model:
\begin{equation}
\label{eq:sv}
\begin{aligned}
\phi_i = \alpha \sum_{\mathcal{S} \subseteq \mathcal{N}\setminus\{t\}} \frac{f(w_{\mathcal{S} \cup \{i\}}) - f(w_\mathcal{S})}{{|\mathcal{N}|-1 \choose |\mathcal{S}|}}.
\end{aligned}
\end{equation}
$\alpha$ is a constant. $\mathcal{S}$ represents the subset of DOs drawn from $\mathcal{N}$. $f(w_\mathcal{S})$ denotes the performance of the FL model $w$ when trained on data owned by $\mathcal{S}$. 
The contributions made by the DOs can be divided into two types: 1) positive contribution (i.e., $\phi_i \geq 0$); and 2) negative contribution (i.e., $\phi_i < 0$). We use the variables $pc_i$ and $nc_i$ to record the number of positive contributions and the number of negative contributions made by each DO $i$, respectively.
Following BRS, the reputation value $v^i$ of $i$ can be computed as follows:
\begin{equation}
\label{eq:reputation}
\begin{aligned}
v^i =\mathbb{E}[Beta(pc_i+1, nc_i +1)] = \frac{pc_i + 1}{pc_i + nc_i + 2}.
\end{aligned}
\end{equation}

It is important to highlight that, as depicted in Eq. \eqref{eq:reputation}, the reputation of each DO $i$ undergoes dynamic updates as the FL model training process unfolds. Furthermore, in cases where there is no prior information available, the default initialization for the reputation value of $i$ is set to the uniform distribution, denoted as $v^i = N(0, 1) = Beta(1, 1)$.

\subsection{Reinforcement Learning Basics}
A Markov Decision Process (MDP) is a mathematical framework for modeling decision-making in which an agent interacts with an environment through discrete time steps. MDP is formally defined by the tuple $\langle S, A, P, R, \gamma \rangle$
\begin{enumerate}
    \item $S$ represents the possible states in the environment, denoted as $s \in S$.
    \item $A$ encompasses the feasible actions the agent can take.
    \item $P: S \times A \times S \rightarrow [0, 1]$ is the transition probability function for the likelihood of transitioning between states when an action is taken, capturing environmental dynamics.
    \item $R: S \times A \times S \rightarrow \mathbb{R}$ is the reward function, specifying immediate rewards upon state transitions due to specific actions, with the agent's aim to maximize cumulative rewards.
    \item $\gamma \in [0, 1]$ serves as the discount factor, reflecting the agent's preference for immediate rewards versus future rewards.
\end{enumerate}
During the MDP process, the agent interacts with the environment across discrete time steps. At each time step, it selects an action $a \in A$ based on policy $\pi : S \rightarrow A$, subsequently receiving a reward $r$, and the environment undergoes state transitions according to $P$.

The goal of MDP is to identify an optimal policy $\pi : S \rightarrow A$ that maximizes the expected sum of discounted rewards over time, given by $\max_{\pi} \mathbb{E} \left[ \sum_{t=1}^T \gamma^{t-1} r^t \right]$. This entails finding the policy maximizing expected cumulative rewards. The value function $V^\pi : S \rightarrow \mathbb{R}$ is associated with each policy, quantifying expected cumulative rewards. The optimal value function $V^* : S \rightarrow \mathbb{R}$ represents the maximum achievable expected cumulative reward achievable with the best policy from each state.

\section{The Proposed \methodname{} Approach}
Our primary objective is to help MUs recruit DOs across multiple sessions while adhering to budget constraints, with the overarching goal of maximizing the total utility. To accomplish this, we must tackle two fundamental challenges:
\begin{enumerate}
    \item \textbf{Budget Allocation}: Determining the allocation of the total budget $B$ to a given session $s$, $B_s$;
    \item \textbf{Bidding Strategy}: Determining the bid price $b_s^i$ for any given DO $i$ in session $s$ under the session budget $B_s$.
\end{enumerate}
Since the AFL market is highly dynamic, it is difficult for MUs to obtain a closed-form analytical solution for the above two problems. Therefore, we design \methodname{} based on reinforcement learning \cite{sutton2018reinforcement} to solve these problems without requiring prior knowledge. 
\begin{figure*}[t!]
\centering
\includegraphics[width=2\columnwidth]{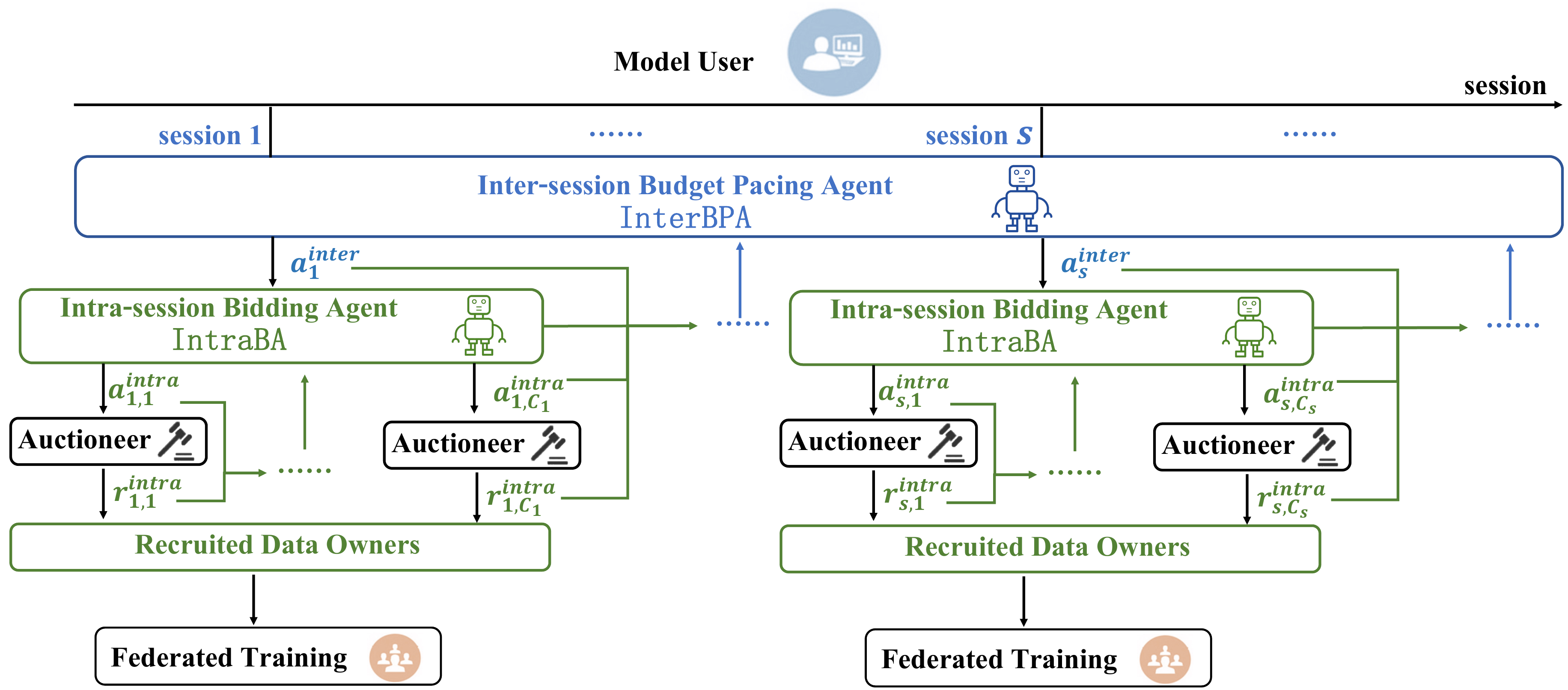}
\caption{An overview of the proposed \methodname{} approach.}
\label{fig:model}
\end{figure*}

To determine the optimal budget allocation strategy and bidding strategy for an MU to realize the objective outlined in Eq. \eqref{eq:goal}, we design \methodname{} based on HRL \cite{HRL2021}. It consists of two HRL-based budget allocation agents: 1) Inter-session Budget Pacing Agent (\interBPA{}), and 2) Intra-session Bidding Agent (\intraBA{}). An overview of \methodname{} is shown in Figure \ref{fig:model}.

During each FL training session $s$, the \interBPA{} observes the current state within the FL model training environment. Subsequently, this observed state is channeled into the policy network of the \interBPA{}, generating the recommended inter-session action (i.e., setting the budget $B_s$ for session $s$). This action aims to enhance the current FL model performance, ultimately influencing the outcome across all training sessions. Moreover, this inter-session action serves as an initial state for the \intraBA{}. It is worth noting that the \interBPA{} will stay static throughout a given session $s$. It is only updated when the session $s$ is concluded. Funneling the inter-session action $B_s$ into the policy network of the \intraBA{} helps determine the intra-session actions, especially the initial intra-session action.

The primary function of the \intraBA{} is to help an MU bid for each DO $i \in [C_s]$ in session $s$ in an efficient way, thus contributing to the crafting of the optimal budget allocation strategies under \methodname{}. The \intraBA{} takes the dynamic MU state as the input, and produces the optimal action $a_s^i$ as the bid price for data owner $i$ to be submitted to the auctioneer. As a result, the \intraBA{} will be updated upon every DO auction in session $s$. The synthesis of inter-session and intra-session actions culminates in the formulation of the MU's budget allocation strategy.
In the following sections, we provide detailed descriptions of these two agents.

\subsection{Inter-session Budget Pacing Agent (\interBPA{})}

\textbf{State}: The state of the \interBPA{} in session $c \in [S]$, denoted as $\boldsymbol{s}_s^{inter}$, comprises two main segments. The first segment contains historical data derived from the preceding $S'$ sessions. These include the budgets allocated for each of the historical sessions, and the bidding outcomes of \intraBA{} in these sessions (including the bid prices for DOs, payment for DOs, and reputation of the recruited DOs). The second segment contains current session information (including the number of available DOs, and the remaining budget). Thus, the formulation of $\boldsymbol{s}_s^{inter}$ is as follows:
\begin{equation}
\label{eq:session_level_state}
\begin{aligned}
\boldsymbol{s}_s^{inter} = \{ & \boldsymbol{b}_{s-S'}, \cdots, \boldsymbol{b}_{s-1}, \boldsymbol{p}_{s-S'}, \cdots, \boldsymbol{p}_{s-1},\boldsymbol{v}_{s-S'}, \cdots, \\
& \boldsymbol{v}_{s-1},  C_s, 
 B, s
\}.
\end{aligned}
\end{equation}
$\boldsymbol{b}_{s-1} = \{b_{s-1}^i\}_{t \in [C_{s-1}]}$, $\boldsymbol{p}_{s-1} = \{p_{s-1}^i\}_{i \in [C_{s-1}]}$, and $\boldsymbol{v}_{s-1} = \{v_{s-1}^i\}_{i \in [C_{s-1}]}$. The integration of historical context into the state design is pivotal, as it empowers the agent to understand the impact of its strategies on FL training over time.

\textbf{Action}: In session $s$, the action to be taken by the \interBPA{} is to determine the budget allocated to the current session, $a_s^{inter}$, which is expressed as:
\begin{equation}
\label{eq:session_level_action}
a_s^{inter} = B_s.
\end{equation}
In this context, $B_s$ denotes the budget designated for session $s$ for bidding for the data owners involved. This inter-session action plays a pivotal role in regulating the amount of budget to be disbursed by the MU during session $s$, thereby helping preserve the total budget $B$ for potential future FL training sessions.

\textbf{Reward}: The inter-session reward for session $s$, $r_s^{inter}$, is determined by the average reputation of DOs recruited in session $s$:
\begin{equation}
\label{eq:session_level_reward}
\begin{aligned}
r_s^{inter} = \frac{1}{\sum_{i \in [C_s]} x_s^i}\sum_{i \in [C_s]} x_s^i v_s^i.
\end{aligned}
\end{equation}
$x_s^i \in \{0,1\}$ denotes if the MU wins the auction for DO $i$. 

\textbf{Discount factor}: As the goal of an MU is to maximize the total utility derived from the recruited DOs for a given total budget $B$ regardless of time, the reward discount factor of \interBPA{} is set as $1$. 

\subsection{Intra-session Budget Management Agent (\intraBA{})}
\textbf{State}: The state of the \intraBA{} in session $s$ during an auction for DO $i$, denoted as $\boldsymbol{s}_{c,i}^{intra}$, consists of: 1) $C_s - i$: the remaining DOs in session $s$, 2) $B_s$: the remaining budget of session $s$, and 3) $v_s^i$: the reputation of DO $i$:
\begin{equation}
\label{eq:data_provider_level_state}
\boldsymbol{s}_{s,i}^{intra} = \{C_s - i, B_s, v_s^i\}. 
\end{equation}

\textbf{Action}: The action, denoted as $\boldsymbol{a}_{s,i}^{intra}$, to be taken by the \intraBA{} in session $s$ for DO $i \in [C_s]$ is to determine the bid price for $i$, i.e., $b_{s}^i$.

\textbf{Reward}: The intra-session reward for session $s$ following the bid for DO $i$ is defined as the utility obtained from $i$, which is formulated as:
\begin{equation}
\label{eq:data_provider_level_reward}
\begin{aligned}
r_{s,i}^{intra} =  x_s^i v_s^i. 
\end{aligned}
\end{equation}

\textbf{Discount factor}: Similar to \interBPA{}, the discount factor for the \intraBA{} is also set as $1$.
\begin{algorithm}[!b]
\caption{The training procedure of \methodname{}}
         Initialize $Q^{intra}$, $Q^{inter}$ with parameters $\theta^{intra}$, $\theta^{inter}$; the target networks of $Q^{intra}$ and $Q^{inter}$ with parameters $\hat{\theta}^{intra}$ and $\hat{\theta}^{inter}$; replay memories $\mathcal{D}^{intra}$ and $\mathcal{D}^{inter}$; the update frequency of target networks, $\Gamma$.
\begin{algorithmic}[1]
\label{alg:training_procedure}
\FOR{$s \in [S]$}
\STATE{Observe state $s_s^{inter}$};
\STATE{Compute $B_s$ according to $\epsilon$-greedy policy w.r.t $Q^{inter}$};
\FOR{$i \in [C_s]$}
\STATE{Observe state $s_{s,i}^{intra}$};
\STATE{Compute $b_s^i$ according to $\epsilon$-greedy policy w.r.t $Q^{intra}$};
\STATE{Submit $b_s^i$ to the auctioneer};
\STATE{Obtain rewards $v_s^i$ and the payment $p_s^i$};
\STATE{$B_s \gets B_s - p_s^i$};
\STATE{Store transition tuples in $\mathcal{D}^{intra}$};
\STATE{Sample a random minibatch of $m$ samples from $\mathcal{D}$};
\STATE{$y^{intra} = r_s^i + \gamma \max_{a_s^{intra'}}Q^{intra}(s_{s,i+1}^{intra},a_s^{intra'};\hat{\theta}^{intra})$};
\STATE{Update $\theta^{intra}$ by minimizing $\sum_{m}[(y^{intra} - Q^{intra}(s_{s,i}^{intra},a_{s,i}^{intra};\theta^{intra})^2]$};
\STATE{$\hat{\theta}^{intra} \gets \theta^{intra}$ every $\Gamma$ steps};
\ENDFOR
\STATE{Obtain rewards $r_s^{inter}$ and the total payment $p_s^i$ during session $s$};
\STATE{$B \gets B - \sum_{i \in [C_s]}p_s^i$};
\STATE{Store transition tuples in $\mathcal{D}^{inter}$};
\STATE{Sample a random minibatch of $m$ samples from $\mathcal{D}$};
\STATE{$y^{inter} = r_s + \gamma \max_{a_s^{inter'}}Q^{inter}(s_{s+1}^{inter},a_s^{inter'};\hat{\theta}^{inter})$};
\STATE{Update $\theta^{inter}$ by minimizing $\sum_{m}[(y^{inter} - Q^{inter}(s_s^{inter},a_s^{inter};\theta^{inter})^2]$};
\STATE{$\hat{\theta}^{inter} \gets \theta^{inter}$ every $\Gamma$ steps};
\ENDFOR
\end{algorithmic}
\end{algorithm}

\subsection{Training Procedure for \interBPA{} and \intraBA{}}
\interBPA{} and \intraBA{} are built on top of the Deep Q-Network (DQN) technique \cite{mnih2015human}. 
A deep neural network (DNN) is adopted to model the action-value function $Q(s, a)$ of both agents, parameterized by $\theta^{inter}$ and $\theta^{intra}$, respectively. To improve stability during training, we pair these networks with a similar DNN architecture parameterized by $\hat{\theta}^{inter}$ and $\hat{\theta}^{intra}$, respectively (referred to as the \textit{target networks}), which also approximates $Q(s, a)$. To update $\theta^{inter}$ and $\theta^{intra}$, the training is conducted by minimizing the following loss function:
\begin{equation}
\label{eq:loss_function}
\begin{aligned}
\mathcal{L}(\theta)=\frac{1}{2}\mathbb{E}_{(s, a, r, s')\sim \mathcal{D}}[(y - Q(s, a;\theta))^2].
\end{aligned}
\end{equation}
The \textit{replay buffer}, $\mathcal{D}$, is a storage mechanism for transition tuples $\{(s, a, r, s')\}_{i=1}^n$, where $s'$ is the new observation following action $a$ based on the state $s$, resulting in reward $r$. This buffer allows the agent to learn from its past experiences by randomly sampling batches of transitions during training.

In the loss function defined in Eq. \eqref{eq:loss_function}, $y$ represents the temporal difference target, and is computed as $y = r + \gamma \max_{a'}Q(s, a';\hat{\theta})$. $\gamma$ is the discount factor, $\hat{\theta}$ represents the parameters of the target network associated with the corresponding agent. $Q(s, a';\hat{\theta})$ is the predicted action-value function of the corresponding agent for its next state $s'$ and all possible actions $a'$. This target network is used to stabilize the learning process by providing a fixed target during training, which is updated periodically (every $\Gamma$ steps) to match the current action-value network.
Algorithm \ref{alg:training_procedure} illustrates the training procedure for \methodname{}.

\section{Experimental Evaluation}
In this section, we evaluate the performance of \methodname{} against seven state-of-the-art AFL approaches based on six real-world datasets.

\subsection{Experiment Settings}
\subsubsection{Dataset}
\label{sec:experiment_setup}
The performance assessment of \methodname{} is conducted on the following six widely-adopted datasets in federated learning studies: 1) MNIST\footnote{http://yann.lecun.com/exdb/mnist/}, 2) CIFAR-10\footnote{https://www.cs.toronto.edu/kriz/cifar.html}, 3) Fashion-MNIST (i.e., FMNIST) \cite{xiao2017fashion}, 4) EMNIST-digits (i.e., EMNISTD), 5) EMNIST-letters (i.e., EMNISTL) \cite{cohen2017emnist} and 6) Kuzushiji-MNIST (i.e., KMNIST) \cite{clanuwat2018deep}.
Similar to \cite{zhang2021incentive}, MNIST, FMNIST, EMNISTD and KMNIST tasks are configured using a base model consisting of an input layer with 784 nodes, a hidden layer with 50 nodes, and an output layer with 10 nodes. However, for EMNISTL tasks, the base model shares the same structure as the aforementioned network, but with an output layer of 26 nodes. Regarding CIFAR-10 tasks, we utilize the streamlined VGG11 network \cite{simonyan2014very}. This network architecture comprises convolutional filters and hidden fully-connected layer sizes set as $\{32, 64, 128, 128, 128, 128, 128, 128\}$ and $128$, respectively.

\subsubsection{Comparison Approaches}
We evaluate the performance of \methodname{} against the following seven AFL bidding approaches in our experiments:
\begin{enumerate}
   \item \textbf{Constant Bid (Const)} \cite{zhang2014optimal}: An MU presents the same bid for all DOs, whereas the bids offered by different MUs can vary.
    \item \textbf{Randomly Generated Bid (Rand)} \cite{zhang2021incentive,zhang2022online}: This approach, commonly found in AFL, involves MUs randomly generating bids from a predefined range for each bid request.
    \item \textbf{Below Max Utility Bid (Bmub)}: This approach is derived from the concept of bidding below max eCPC \cite{lee2018estimating} in online advertisement auctioning. It defines the utility of each bid request from a DO as the upper limit of the bid values offered by MUs. Therefore, for each bid request, the bid price is randomly generated within the range between 0 and this upper bound.
    \item \textbf{Linear-Form Bid (Lin)} \cite{perlich2012bid}: This strategy generates bid values which are directly proportional to the estimated utility of the bid requests, typically expressed as $b^{Lin} (v^i)= \lambda_{Lin} v^i$.
    \item \textbf{Bidding Machine (BM)} \cite{ren2017bidding}: Commonly used in online advertisement auctioning, especially in real-time bidding, this method focuses on maximizing a specific buyer's profit by optimizing outcome prediction, cost estimation, and the bidding strategy.
    \item \textbf{Fed-Bidder} \cite{tang2023utility}: This bidding method is specifically designed for MUs in AFL settings. It guides them to competitively bid for DOs to maximize their utility. It has two variants, one with a simple winning function, referred to as Fed-Bidder-sim \textbf{(FBs)}; and the other with a complex winning function, referred to as Fed-Bidder-com \textbf{(FBc)}.
    \item \textbf{Reinforcement Learning-based Bid (RLB)} \cite{cai2017real}: It regards the bidding process as a reinforcement learning problem, utilizing an MDP framework to learn the most effective bidding policy for an individual buyer to enhance the auctioning outcomes.
\end{enumerate}

\subsubsection{Experiment Scenarios}
We compare the performance of the proposed \methodname{} with baseline methods under three experiment scenarios with each containing 10,000 DOs:
\begin{enumerate}
    \item \textbf{IID data, varying dataset sizes, without noise}: In this scenario, the sizes of datasets owned by various DOs are randomly generated, ranging from 500 to 5,000 samples. Additionally, all the data are independent and identically distributed (IID), with no noise.

    \item \textbf{IID data, same dataset size, with noise}: Each DO shares the same number of data samples (i.e., 3,000 images) including noisy ones. 
    In particular, we categorize the 10,000 DOs into 5 sets, each comprising 2,000 DOs. Then, we introduce varying amounts of noisy data for each set of DOs, as follows:
    The first set of DOs contains 0\% noisy data.
    The second set of DOs includes 10\% noisy data.
    The third set of DOs involves 25\% noisy data.
    The fourth set of DOs consists of 40\% noisy data.
    The last set of DOs comprises 60\% noisy data.

    \item \textbf{Non-IID data, with noise}: In this experimental scenario, we deliberately introduce data heterogeneity by adjusting the class distribution among individual DOs. Following the methodology outlined in \cite{shi2023fairfed}, we implement the following Non-IID setup. We designate one class (on MNIST, CIFAR, FMNIST, EMNISTD, and KMNIST) or six classes (on EMNISTL) as the minority class and assign this minority class to 100 DOs. As a result, these 100 DOs possess images for all classes, while all other DOs exclusively have images for the remaining nine classes, excluding the minority class. In this experiment scenario, each DO holds 3,000 images. Additionally, we simulate scenarios in which the minority DOs contain 10\% or 25\% noisy data.
\end{enumerate}

\subsubsection{Implementation Details}
In our experiments, we faced the challenge of not having a publicly available AFL bidding behaviour dataset. To address this issue, we track the behaviors of MUs over time during simulations to gradually accumulate data in four different settings. Each setting contains 160 MUs who adopted one of the eight bidding strategies listed in the Compared Approaches section.

In the first setting, each of the eight baseline bidding methods is adopted by one eighth of the MUs. In the second setting, as BM, Fed-Bidder variants (FBs and FBc) and RLB have AI techniques similar to \methodname{}, these four bidding strategies are adopted by three sixteenths of the total population, while the remaining four baselines are adopted by one sixteenth of the total population.
In the third and fourth settings, as both Fed-Bidder variants and \methodname{} are designed specifically for AFL, we set the percentage of MUs adopting FBs and FBc to be higher than those adopting the other six baselines. Specifically, under the third setting, 50 MUs adopt FBs and FBc, while 10 MUs adopt each of the other six baselines. Under the fourth setting, 65 MUs adopted FBs and FBc, while 5 MUs adopted each of the other six baselines.
We adopt the generalized second-price sealed-bid forward auction (GSP) mechanism in our experiments. By tracking the behaviors of MUs over time, we can gradually accumulate data in the absence of a publicly available dataset related to AFL bidding behaviours.

To evaluate the effectiveness of \methodname{}, we create nine MUs, each utilizing one of the aforementioned bidding approaches to join the auction for each bid request (i.e., each DO) in each session $s$. Following \cite{tang2023utility}, bid requests are delivered in chronological order. Upon receiving a bid request, each MU derives its bid price based on its adopted bidding strategy.
Subsequently, the auctioneer gathers the bid prices, identifies the winner, and determines the market price using the GSP auction mechanism. The winning MU pays the market price to the DO.
The process concludes when there are no more bid requests or when the budget is depleted.

\methodname{} utilizes fully connected neural networks with three hidden layers each containing 64 nodes to generate bid prices for a target DO on behalf of their respective MUs. The replay buffer $\mathcal{D}$ of both the \interBPA{} and the \intraBA{} are set to 5,000. During training, both agents explore the environment using an $\epsilon$-greedy policy with an annealing rate from 1.0 to 0.05. In updating both $Q^{intra}$ and $Q^{inter}$, 64 tuples uniformly sampled from $\mathcal{D}$ are used for each training step, and the corresponding target networks are updated once every 20 steps. In our experiments, we use RMSprop with a learning rate of 0.0005 to train all neural networks, and set the discount factor $\gamma$ to 1. In addition, we have set the number of candidate DOs within each session to 200 (i.e., $C_s = 200$).
The communication round in each session is set at 100, while the local training epochs is set at 30. 
The detailed hyperparameter settings are shown in Table \ref{tab:parameter_setting}. 

\begin{table}[!t]
\centering
\caption{Experiment settings.} 
\resizebox*{0.52\linewidth}{!}{
\begin{tabular}{|*{2}{c|}}
\hline
Parameter & Setting\\\hline
Batch size & 512\\
Local training epochs & 30 \\
$C_s$ & 200 \\
$S$ & 100 \\
$\eta $ & 0.0005  \\
$\mathcal{D}^{intra}$,  $\mathcal{D}^{inter}$ & 5,000 \\
$\Gamma$ & 20 \\
$\gamma$ & 1 \\
$m$ & 64  \\\hline
\end{tabular}
}
\label{tab:parameter_setting}
\end{table}

\begin{table*}[ht]
\centering
\caption{Comparison of total number of data samples obtained and utilities across different budget settings and datasets, under the scenario of IID data, different sizes of DOs datasets without noisy samples.} 
\resizebox*{1\linewidth}{!}{
\begin{tabular}{|*{14}{c|}}
\hline
\multirow{2}*{Budget} & \multirow{2}*{Method} & \multicolumn{2}{c|}{MNIST} & \multicolumn{2}{c|}{CIFAR} &\multicolumn{2}{c|}{FMNIST}& \multicolumn{2}{c|}{EMNIST} & \multicolumn{2}{c|}{EMNISTL} & \multicolumn{2}{c|}{KMNIST}\\\cline{3-14}
{} & {} & \#data & utility & \#data & utility & \#data & utility & \#data & utility & \#data & utility & \#data & utility\\\hline
\multirow{9}*{100}  & Const & 8,832 & 7.36  & 9,897 & 7.87 & 10,722 & 6.46 & 7,638 & 6.52 & 7,359 & 7.02 & 7,810 & 6.75\\
{} & Rand & 9,125 & 8.41  & 8,721 & 8.43 & 9,743 & 8.09 & 8,853 & 8.10 & 6,822 & 7.97 & 8,940 & 7.96\\
{} & Bmub & 9,246 & 9.03  & 11,302 & 9.19 & 12,274 & 8.76 & 10,382 & 8.91 & 6,485 & 9.15 & 10,551 & 8.62\\
{} & Lin & 9,461 & 10.28  & 11,426 & 10.17 & 13,523 & 9.84 & 10,673 & 10.33 & 8,220 & 10.51 & 10,694 & 9.97\\
{} & BM & 12,324 & 11.95  & 13,367 & 11.85 & 15,321 & 12.65 & 14,399 & 12.19 & 15,157 & 12.27 & 14,501 & 12.46\\
{} & FBs & \underline{13,985} & \underline{14.51}  & 14,259 & 13.51 & 16,373 & 13.53 & 15,321 & 13.46 & 14,408 & 13.44 & 15,509 & 13.54\\
{} & FBc & 13,869 & 13.84  & 13,984 & 13.70 & 15,843 & 13.42 & \underline{16,772} & \underline{14.23} & 14,168 & 13.67 & \underline{16,927} & 13.64\\
{} & RLB & 13,892 & 14.42  & \underline{14,263} & \underline{14.26} & \underline{17,783} & \underline{13.95} & 15,989 & 13.51 & \underline{15,544} & \underline{14.40} & 16,027 & \underline{14.33}\\\cline{2-14}
{} & \methodname{} & \textbf{14,944} & \textbf{16.59}  & \textbf{17,397} & \textbf{17.47} & \textbf{19,064} & \textbf{18.19} & \textbf{18,674} & \textbf{17.46} & \textbf{16,317} & \textbf{18.59} & \textbf{18,687} & \textbf{16.55}\\\hline
\multirow{9}*{200}  & Const & 11,037 & 8.49  & 12,043 & 9.31 & 16,374 & 8.52 & 13,826 & 9.46 & 10,876 & 10.33 & 13,950 & 9.31\\
{} & Rand & 10,895 & 10.06  & 11,894 & 10.00 & 14,898 & 9.90 & 12,452 & 10.34 & 12,808 & 10.42 & 12,601 & 10.05\\
{} & Bmub & 16,582 & 9.58  & 17,021 & 10.60 & 25,327 & 10.60 & 17,817 & 10.40 & 20,966 & 11.43 & 17,878 & 10.97\\
{} & Lin & 17,803 & 13.14  & 17,849 & 12.88 & 26,880 & 12.88 & 19,435 & 12.64 & 27,860 & 12.70 & 19,553 & 12.97\\
{} & BM & 23,584 & 14.97  & 20,836 & 15.11 & 31,945 & 15.92 & 21,656 & 15.03 & 35,016 & 15.29 & 21,722 & 15.70\\
{} & FBs & 27,813 & 17.70  & 28,456 & 17.61 & 34,936 & 17.09 & 26,994 & 17.01 & 31,743 & 17.40 & 27,087 & 17.49\\
{} & FBc & 28,005 & 17.51  & 29,835 & 17.24 & \underline{36,873} & \underline{17.58} & \underline{27,863} & 16.60 & 34,686 & 16.99 & \underline{27,892} & \underline{17.89}\\
{} & RLB & \underline{29,468} & \underline{17.77}  & \underline{30,138} & \underline{17.82} & 35,548 & 17.04 & 26,748 & \underline{17.45} & \underline{37,122} & \underline{17.82} & 26,819 & 17.23\\\cline{2-14}
{} &\methodname{} & \textbf{33,045} & \textbf{21.99}  & \textbf{35,163} & \textbf{21.08} & \textbf{39,982} & \textbf{23.72} & \textbf{35,656} & \textbf{19.59} & \textbf{37,645} & \textbf{22.43} & \textbf{35,737} & \textbf{18.08}\\\hline
\multirow{9}*{400}  & Const & 14,395 & 8.72  & 15,362 & 8.11 & 18,475 & 8.34 & 17,877 & 7.82 & 10,177 & 8.04 & 17,940 & 8.41\\
{} &Rand & 13,195 & 9.86  & 16,372 & 9.71 & 17,844 & 6.87 & 17,003 & 7.13 & 6,431 & 9.02 & 17,051 & 9.20\\
{} &Bmub & 23,378 & 10.90  & 25,631 & 11.16 & 31,487 & 10.86 & 24,756 & 10.05 & 23,639 & 10.63 & 24,869 & 11.33\\
{} &Lin & 24,523 & 14.58  & 26,830 & 14.41 & 32,677 & 14.24 & 25,669 & 14.28 & 36,261 & 14.31 & 25,802 & 14.46\\
{} &BM & 38,516 & 16.46  & 30,173 & 16.54 & 38,552 & 16.90 & 30,878 & 17.26 & 41,050 & 17.66 & 31,077 & 17.61\\
{} &FBs & 50,983 & 19.32  & 38,452 & 19.24 & 39,236 & 18.54 & 38,452 & 18.69 & 40,605 & 19.04 & 38,566 & 19.09\\
{} &FBc & 50,146 & 19.23  & 39,817 & 19.10 & 41,582 & 18.37 & \underline{40,663} & 18.40 & 39,555 & 18.85 & \underline{40,768} & 18.88\\
{} &RLB & \underline{51,643} & \underline{19.54}  & \underline{42,731} & \underline{19.63} & \underline{45,667} & \underline{18.84} & 37,748 & \underline{19.18} & \underline{43,077} & \underline{19.71} & 37,843 & \underline{19.55}\\\cline{2-14}
{} &\methodname{} & \textbf{56,872} & \textbf{23.65}  & \textbf{53,672} & \textbf{22.71} & \textbf{52,386} & \textbf{23.00} & \textbf{47,135} & \textbf{19.32} & \textbf{46,341} & \textbf{23.83} & \textbf{47,262} & \textbf{19.73}\\\hline
\multirow{9}*{600}  & Const & 17,895 & 9.71  & 19,378 & 9.60 & 21,394 & 9.33 & 19,832 & 10.08 & 10,596 & 9.55 & 19,982 & 8.92\\
{} &Rand & 19,803 & 8.68  & 20,184 & 9.07 & 20,853 & 11.69 & 18,838 & 10.37 & 24,581 & 9.15 & 18,966 & 9.83\\
{} &Bmub & 30,164 & 12.07  & 29,174 & 11.93 & 37,421 & 11.85 & 29,669 & 12.06 & 33,768 & 11.94 & 29,845 & 11.97\\
{} &Lin & 32,973 & 15.62  & 30,375 & 15.59 & 40,128 & 15.08 & 34,452 & 15.16 & 47,484 & 15.61 & 34,629 & 15.62\\
{} &BM & 49,807 & 17.09  & 49,272 & 17.43 & 47,533 & 18.06 & 38,743 & 17.85 & 51,454 & 18.23 & 38,943 & 18.54\\
{} &FBs & 62,396 & 20.49  & 50,384 & 20.58 & 46,731 & 19.54 & 45,232 & 19.64 & 50,482 & 20.29 & 45,288 & 20.29\\
{} &FBc & 61,478 & 20.31  & 52,836 & 20.24 & \underline{52,843} & \underline{19.92} & \underline{48,767} & 19.38 & 49,468 & 20.04 & \underline{48,958} & 20.06\\
{} &RLB & \underline{63,672} & \underline{20.64}  & \underline{58,273} & \underline{20.64} & 50,472 & 19.26 & 42,534 & \underline{19.69} & \underline{59,455} & \underline{20.53} & 42,692 & \underline{20.44}\\\cline{2-14}
{} &\methodname{} & \textbf{66,654} & \textbf{21.72} & \textbf{60,737} & \textbf{22.82} & \textbf{63,824} & \textbf{24.17} & \textbf{58,462} & \textbf{23.01} & \textbf{63,441} & \textbf{23.54} & \textbf{58,522} & \textbf{21.72}\\\hline 
\multirow{9}*{800}  & Const & 23,047 & 11.04  & 24,753 & 11.35 & 26,311 & 11.13 & 22,644 & 10.79 & 17,875 & 11.40 & 22,705 & 11.30\\
{} &Rand & 24,853 & 14.09  & 22,845 & 13.34 & 22,734 & 13.68 & 20,474 & 13.60 & 26,563 & 13.57 & 20,642 & 13.26\\
{} &Bmub & 36,703 & 12.99  & 35,777 & 12.70 & 40,275 & 13.47 & 36,648 & 12.91 & 38,570 & 13.08 & 36,732 & 13.17\\
{} &Lin & 39,651 & 16.79  & 38,561 & 16.88 & 47,823 & 16.55 & 40,537 & 16.67 & 59,390 & 16.86 & 40,727 & 16.76\\
{} &BM & 57,442 & 18.57  & 52,735 & 18.68 & 51,272 & 19.16 & 46,772 & 19.34 & 65,086 & 19.41 & 46,933 & 19.59\\
{} &FBs & 70,496 & 22.09  & 62,842 & 22.07 & 54,453 & 21.07 & 51,863 & 21.02 & 67,470 & 21.54 & 51,942 & 21.69\\
{} &FBc & \underline{72,845} & 22.04  & 63,112 & 22.06 & \underline{55,388} & \underline{21.18} & \underline{56,991} & \underline{21.09} & 61,598 & 21.57 & 57,152 & 21.53\\
{} &RLB & 70,381 & \underline{22.31}  & \underline{66,843} & \underline{22.37} & 52,621 & 20.92 & 53,823 & 20.95 & \underline{68,943} & \underline{21.78} & \underline{57,900} & \underline{21.92}\\\cline{2-14}
{} &\methodname{} & \textbf{77,821} & \textbf{22.40}  & \textbf{71,244} & \textbf{23.46} & \textbf{64,739} & \textbf{23.12} & \textbf{62,579} & \textbf{22.57} & \textbf{70,393} & \textbf{23.04} & \textbf{59,711} & \textbf{22.18}\\\hline
\end{tabular}
}
\label{tab:bidding_performance}
\end{table*}

\begin{figure*}[t!] 
\centering
\includegraphics[width=1\linewidth
, height=0.56\linewidth
]{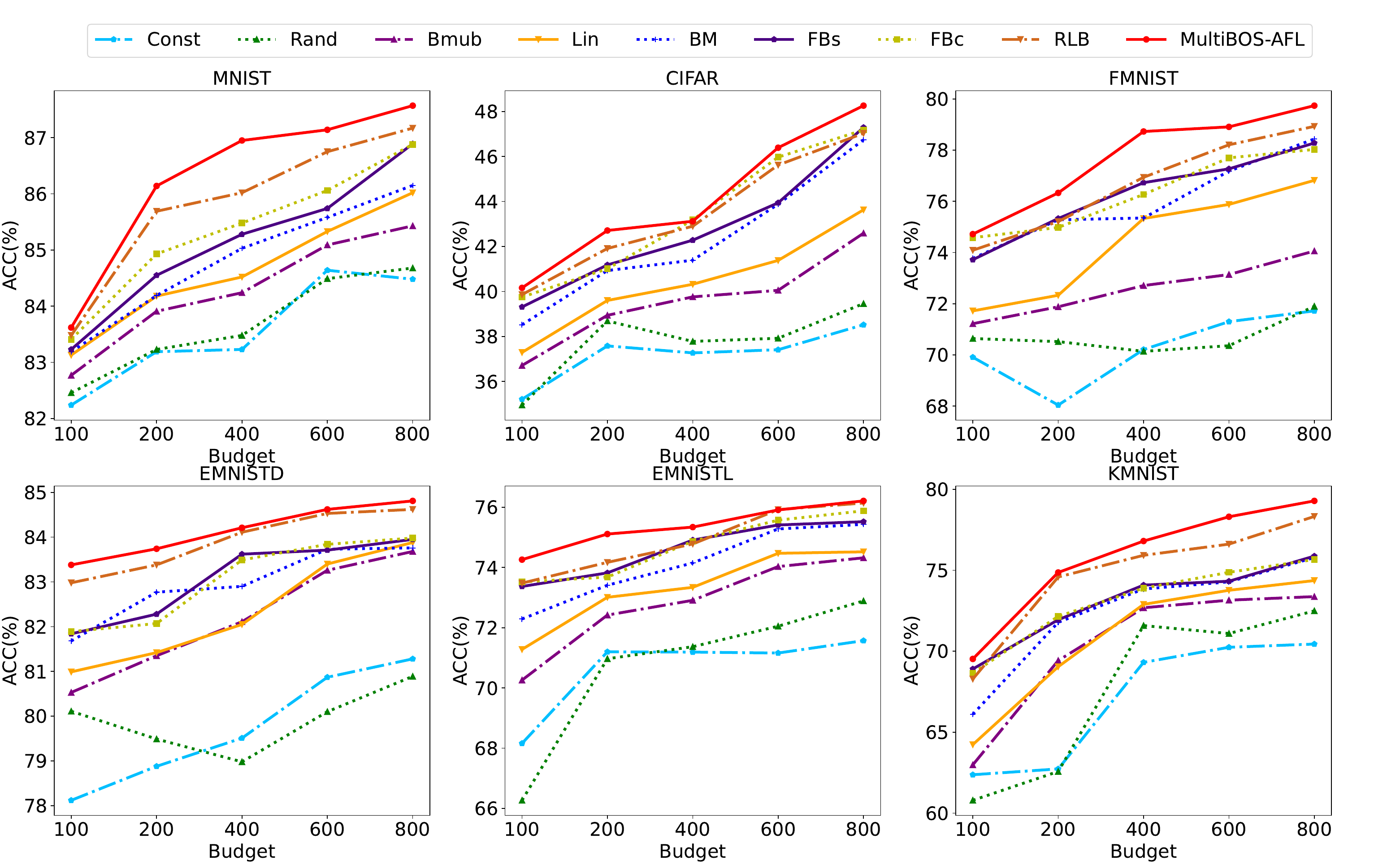}
\caption{Comparison of test accuracies achieved by the FL models produced by different approaches (DO datasets are of different sizes and without noisy sample).}
\label{fig:acc_different_datasamples}
\end{figure*}

\begin{figure*}[t!] 
\centering
\includegraphics[width=1\linewidth
, height=0.56\linewidth
]{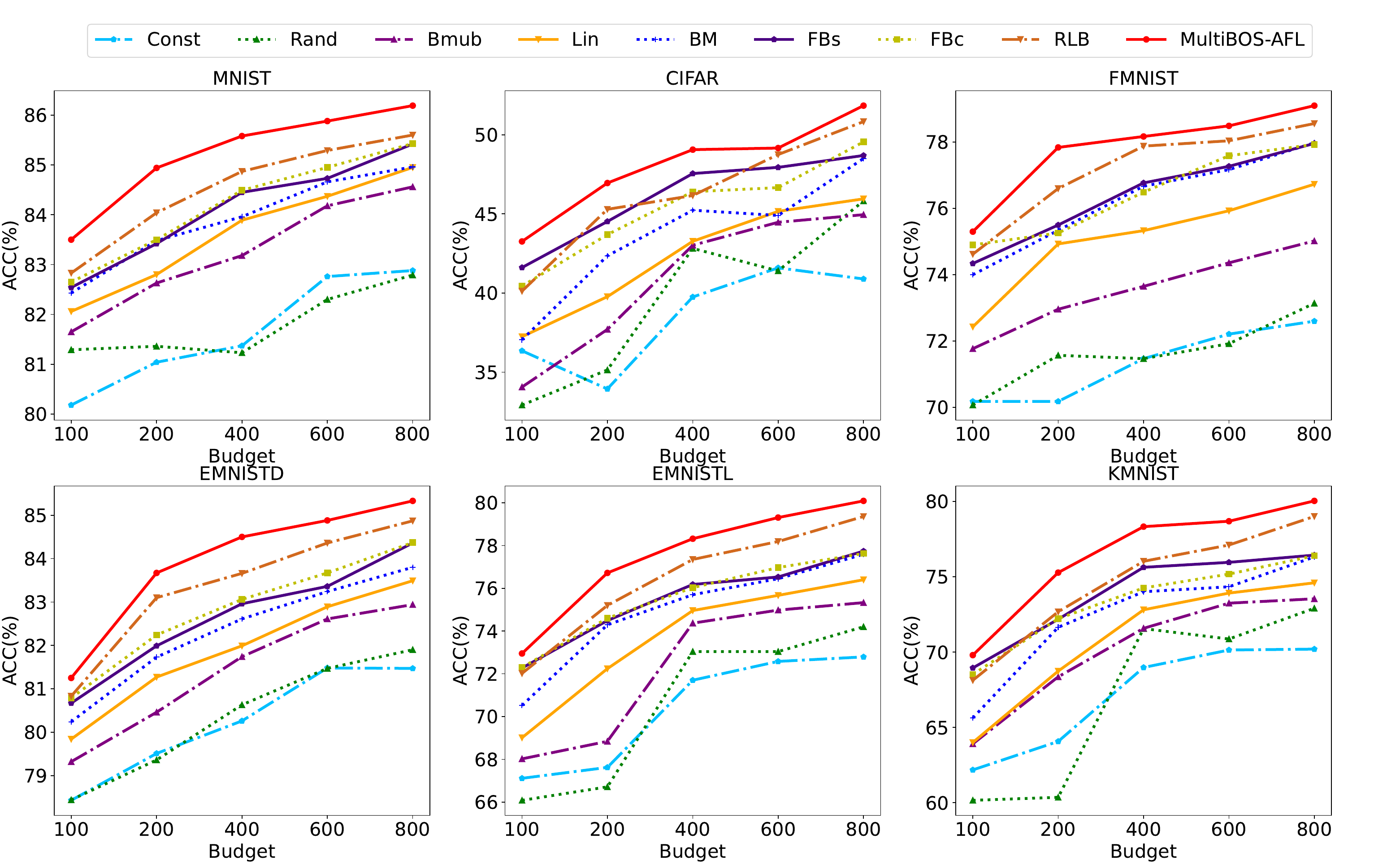}
\caption{Comparison of test accuracies achieved by the FL models produced by different approaches (DO datasets are of the same size and with noisy samples).}
\label{fig:acc_different_noisy_data_percentage}
\end{figure*}

\subsection{Evaluation Metrics}
To evaluate the effectiveness of all the comparison methods, we adopt the following three metrics:
\begin{itemize}
    \item The number of data samples won by the model user (\textbf{\#data}) is defined as the cumulative number of data samples owned by all DOs recruited by the corresponding model user until the budget or session limits are reached.
    \item The utility obtained by the model user (\textbf{utility}) is defined as the cumulative reputation of DOs recruited by the corresponding model user until the budget or session limits are reached.
    \item The test accuracy (\textbf{Acc}) is determined as the accuracy of the final FL model for the respective model user, up to the point where either the budget or session limits are reached.
\end{itemize}

\subsection{Results and Discussion}
To conduct a comparative analysis of bidding strategies based on these metrics, we carry out experiments across six datasets, each with varying budget settings.These settings span the range of $\{100, 200, 400, 600, 800\}$. The results are shown in Tables \ref{tab:bidding_performance}, \ref{tab:performance_different_noisy_data_percentage}, \ref{tab:acc_non_iid}, and Figures \ref{fig:acc_different_datasamples} and \ref{fig:acc_different_noisy_data_percentage}.

From Table \ref{tab:bidding_performance}, which shows the results of various comparison methods under the IID data, different sizes of DOs datasets without noisy samples scenario, it can be observed that under all six datasets and five budget settings, our proposed \methodname{} approach consistently outperforms all baseline methods in terms of both evaluation metrics. Specifically, compared to the best-performing baseline, \methodname{} achieves 12.28\%  and 14.52\% improvement in terms of total utility and the number of data samples won, respectively. 
Figure \ref{fig:acc_different_datasamples} shows the corresponding test accuracy. The results align with the auction performance shown in Table \ref{tab:bidding_performance} with \methodname{} improving the test accuracy by 1.23\% on average.

Table \ref{tab:performance_different_noisy_data_percentage} and Figure \ref{fig:acc_different_noisy_data_percentage} show the utility obtained by the corresponding model users adopting these nine comparison methods and the accuracy of the FL models, respectively, under the IID data, same sizes of DOs datasets with noisy samples scenario. It can be observed that in this experiment scenario, the results are in consistent with those shown in Table \ref{tab:bidding_performance} and Figure \ref{fig:acc_different_noisy_data_percentage}. The proposed method \methodname{} improves the utility and accuracy of the model obtained by the corresponding MU by 2.41\% and 1.27\% on average, respectively under this scenario. 

In addition, the comparative results under the Non-IID data with noise scenario can be found in Table \ref{tab:acc_non_iid}. It can be observed that under these two different settings, the proposed method \methodname{} consistently outperforms existing methods in terms of achieving higher FL model accuracy. In particular, on average, \methodname{} achieves 1.49\% and 1.72\% higher FL model accuracy compared to the best performance achieved by baselines under the 10\% noisy data and 25\% noisy data settings, respectively.  
All these results demonstrate the effectiveness of our approach in helping MUs optimize their budget pacing and bidding strategies for DOs under the emerging multi-session AFL scenarios.

Among all the comparison methods, Lin and Bmub typically outperform Const and Rand due to their use of utility in the bidding process. However, Bmub is less effective than Lin due to its reliance on randomness. Meanwhile, the more advanced methods BM, FBs, FBc, RLB and \methodname{} perform significantly better than the simpler approaches. This is largely due to the inclusion of auction records (including auction history and bidding records) and the use of advanced learning methods.

RLB and \methodname{} both outperform BM, FBs, and FBc, due to their ability of adaptive adjustment to the highly dynamic auction environment. While BM does consider market price distribution, it derives this distribution by learning the prediction of each bid request's market price density, which may lead to overfitting. In contrast, FBs and FBc obtain the market price distribution via a predefined winning function, which helps predict the expected bid costs more accurately. However, BM, FBs and FB are still static bidding strategies. They are essentially represented by linear or non-linear functions whose parameters are derived from historical auction data using heuristic techniques. Subsequently, these parameters are applied to new auctions, even if the dynamics of these new auctions may vary significantly from those in the historical data. The inherent dynamism of the AFL market poses a considerable challenge for these static bidding methods, making it hard for them to consistently achieve desired outcomes in subsequent auctions.

It is important to note that while RLB employs dynamic programming to optimize its bidding process, it is susceptible to the drawback of immediate reward setting, which might result in indiscriminate bidding for data samples without considering their associated costs. This issue is effectively addressed by \methodname{}. Moreover, it is worth highlighting that RLB is not designed for optimizing budget allocation across multiple sessions. This is a distinction where \methodname{} offers significant advantages.

The test accuracy achieved by the FL models trained under all bidding strategies on CIFAR-10 is consistently lower than that on other datasets. This can be attributed to the base model adopted for FL training. As mentioned in Section \ref{sec:experiment_setup}, the accuracy reported in these two figures is with regard to the VGG11 network. Nevertheless, even with such a less effective base model, \methodname{} still significantly outperforms other baselines.

\begin{table*}[ht]
\centering
\caption{Utility comparison across different budget settings and datasets under the scenario of IID data, same sizes of DOs datasets with noisy samples. } 
\resizebox*{0.66\linewidth}{!}{
\begin{tabular}{|*{8}{c|}}
\hline
Budget & Method & MNIST & CIFAR & FMNIST & EMNIST & EMNISTL & KMNIST\\\hline
\multirow{9}*{100}  & Const &6.94 & 6.04 & 6.95 & 7.51 & 6.82 & 6.70\\ 
{} & Rand & 8.01 & 7.69 & 7.96 & 8.44 & 8.09 & 8.05\\ 
{} & Bmub & 8.66 & 8.38 & 9.00 & 9.17 & 9.03 & 8.71\\ 
{} & Lin & 10.26 & 9.82 & 10.02 & 10.25 & 10.13 & 10.05\\ 
{} & BM & 12.14 & 12.85 & 12.73 & 11.91 & 12.58 & 12.40\\ 
{} & FBs & 13.72 & 13.34 & 13.51 & 13.65 & 13.65 & 13.63\\ 
{} & FBc & 13.77 & 13.47 & 13.68 & 13.71 & 13.69 & 13.65\\ 
{} & RLB & \underline{14.65} & \underline{14.18} & \underline{14.12} & \underline{14.24} & \underline{14.13} & \underline{14.30}\\\cline{2-8}
{} & \methodname{} & \textbf{15.14} & \textbf{14.86} & \textbf{14.32} & \textbf{14.95} & \textbf{14.33} & \textbf{14.81}\\\hline

\multirow{9}*{200}  & Const & 9.53 & 9.56 & 9.39 & 8.88 & 8.94 & 9.02\\ 
{} & Rand & 10.25 & 10.10 & 9.98 & 10.05 & 10.04 & 10.08\\ 
{} & Bmub & 10.51 & 11.53 & 11.64 & 10.07 & 10.84 & 10.56\\ 
{} & Lin & 13.07 & 12.80 & 12.94 & 12.91 & 12.95 & 12.97\\ 
{} & BM & 15.15 & 16.10 & 16.19 & 15.01 & 15.82 & 15.54\\ 
{} & FBs & 17.75 & 17.14 & 17.47 & 17.47 & 17.37 & 17.42\\ 
{} & FBc & 17.36 & 16.89 & 17.42 & 17.19 & 17.32 & 17.20\\ 
{} & RLB & \underline{17.91} & \underline{17.48} & \underline{17.96} & \underline{17.66} & \underline{17.52} & \underline{17.78}\\\cline{2-8}
{} & \methodname{} & \textbf{18.18} & \textbf{18.51} & \textbf{18.14} & \textbf{17.99} & \textbf{17.93} & \textbf{18.25}\\\hline

\multirow{9}*{400}  & Const & 8.55 & 8.17 & 8.55 & 8.23 & 8.57 & 8.45\\ 
{} & Rand & 10.63 & 7.64 & 8.76 & 8.91 & 8.20 & 8.75\\ 
{} & Bmub & 11.19 & 11.15 & 11.44 & 10.75 & 10.96 & 11.03\\ 
{} & Lin & 14.65 & 14.31 & 14.27 & 14.40 & 14.30 & 14.45\\ 
{} & BM & 17.75 & 18.18 & 17.30 & 16.38 & 16.95 & 17.32\\ 
{} & FBs & 19.48 & 18.70 & 18.89 & 19.09 & 18.82 & 19.01\\ 
{} & FBc & 19.27 & 18.41 & 18.82 & 18.95 & 18.74 & 18.82\\ 
{} & RLB & \underline{19.97} & \underline{19.26} & \underline{19.37} & \underline{19.39} & \underline{19.20} & \underline{19.40}\\\cline{2-8}
{} & \methodname{} & \textbf{20.24} & \textbf{19.49} & \textbf{19.51} & \textbf{20.48} & \textbf{19.33} & \textbf{19.57}\\\hline

\multirow{9}*{600}  & Const & 9.13 & 8.75 & 8.80 & 9.89 & 9.29 & 9.23\\ 
{} & Rand & 8.18 & 11.20 & 10.47 & 9.42 & 10.47 & 9.94\\ 
{} & Bmub & 12.14 & 11.92 & 11.72 & 11.98 & 11.83 & 12.00\\ 
{} & Lin & 15.92 & 15.37 & 15.52 & 15.42 & 15.37 & 15.50\\ 
{} & BM & 18.28 & 19.25 & 18.44 & 17.16 & 17.91 & 18.17\\ 
{} & FBs & 20.71 & 19.76 & 20.19 & \underline{20.39} & 19.97 & 20.13\\ 
{} & FBc & 20.57 & 19.52 & 19.91 & 19.98 & 19.73 & 19.92\\ 
{} & RLB & \underline{20.69} & \underline{19.98} & \underline{20.47} & 20.26 & \underline{20.26} & \underline{20.31}\\\cline{2-8}
{} & \methodname{} & \textbf{21.33} & \textbf{20.78} & \textbf{20.71} & \textbf{20.75} & \textbf{20.46} & \textbf{20.55}\\\hline

\multirow{9}*{800}  & Const & 11.15 & 11.24 & 11.74 & 11.10 & 11.40 & 11.14\\ 
{} & Rand & 13.43 & 13.02 & 13.64 & 13.76 & 13.98 & 13.55\\ 
{} & Bmub & 12.90 & 13.39 & 13.63 & 12.85 & 13.55 & 13.19\\ 
{} & Lin & 16.87 & 16.64 & 16.75 & 16.78 & 16.68 & 16.71\\ 
{} & BM & 19.52 & 20.11 & 19.41 & 18.54 & 19.08 & 19.34\\ 
{} & FBs & 22.10 & 21.08 & 21.55 & 21.82 & 21.43 & 21.59\\ 
{} & FBc & 21.97 & \underline{21.24} & 21.54 & 21.80 & 21.41 & 21.44\\
{} & RLB & \underline{22.37} & 20.84 & \underline{21.78} & \underline{22.04} & \underline{21.60} & \underline{21.77}\\\cline{2-8}
{} & \methodname{} & \textbf{24.60} & \textbf{21.62} & \textbf{22.04} & \textbf{22.57} & \textbf{21.82} & \textbf{22.21}\\\hline 
\end{tabular}
}
\label{tab:performance_different_noisy_data_percentage}
\end{table*}

\begin{table*}[ht]
\centering
\caption{FL model accuracy comparison across different budget settings and datasets under the Non-IID data with noise scenario. \textit{ND} represents noisy data.} 
\resizebox*{1\linewidth}{!}{
\begin{tabular}{|*{14}{c|}}
\hline
\multirow{2}*{Budget} & \multirow{2}*{Method} & \multicolumn{2}{c|}{MNIST} & \multicolumn{2}{c|}{CIFAR} &\multicolumn{2}{c|}{FMNIST}& \multicolumn{2}{c|}{EMNIST} & \multicolumn{2}{c|}{EMNISTL} & \multicolumn{2}{c|}{KMNIST}\\\cline{3-14}
{} & {} & 10\% ND & 25\% ND & 10\% ND & 25\% ND & 10\% ND & 25\% ND & 10\% ND & 25\% ND & 10\% ND & 25\% ND & 10\% ND & 25\% ND\\\hline
\multirow{9}*{100}  & Const &70.11 & 70.03 & 12.88 & 13.97 & 61.48 & 57.87 & 77.02 & 76.46 & 64.92 & 63.30 & 58.21 & 59.63\\ 
{} & Rand & 69.61 & 65.42 & 10.57 & 10.83 & 62.70 & 59.48 & 78.69 & 77.97 & 63.97 & 62.83 & 57.01 & 59.12\\ 
{} & Bmub & 71.22 & 70.61 & 15.37 & 12.94 & 63.32 & 60.45 & 78.42 & 77.37 & 66.88 & 65.19 & 61.83 & 61.76\\ 
{} & Lin & 72.36 & 70.32 & 18.65 & 17.41 & 64.04 & 64.13 & 78.62 & 77.44 & 66.47 & 64.07 & 62.72 & 62.97\\ 
{} & BM & 72.31 & 71.65 & 19.50 & 19.62 & 67.35 & 66.25 & 79.50 & 78.42 & 67.17 & 64.62 & 64.55 & 63.77\\ 
{} & FBs & \underline{73.23} & 72.32 & \underline{23.59} & 22.03 & 70.97 & 70.26 & 79.51 & 78.35 & \underline{68.35} & \underline{65.94} & \underline{65.82} & \underline{64.33}\\ 
{} & FBc & 73.11 & \underline{74.80} & 23.42 & 22.26 & \underline{71.29} & \underline{70.68} & \underline{79.92} & \underline{78.93} & 67.69 & 64.78 & 65.47 & 63.88\\ 
{} & RLB & 73.07 & 73.11 & 22.94 & \underline{22.98} & 71.03 & 69.55 & 79.83 & 78.66 & 68.20 & 65.57 & 65.38 & 63.93\\\cline{2-14}
{} & \methodname{} & \textbf{73.79} & \textbf{75.22} & \textbf{23.88} & \textbf{23.24} & \textbf{72.31} & \textbf{71.42} & \textbf{80.66} & \textbf{79.29} & \textbf{69.26} & \textbf{66.76} & \textbf{66.15} & \textbf{65.08}\\\hline

\multirow{9}*{200}  & Const & 70.73 & 66.38 & 10.68 & 11.08 & 63.74 & 60.16 & 77.98 & 77.52 & 67.84 & 66.16 & 58.44 & 58.29\\ 
{} & Rand & 69.48 & 68.96 & 10.32 & 10.26 & 63.86 & 59.63 & 78.63 & 78.19 & 68.24 & 66.88 & 59.25 & 58.09\\ 
{} & Bmub & 71.81 & 70.52 & 13.39 & 13.03 & 63.83 & 62.18 & 79.37 & 78.37 & 69.09 & 67.42 & 63.04 & 63.34\\ 
{} & Lin & 72.98 & 70.55 & 19.07 & 17.96 & 64.43 & 64.16 & 79.43 & 78.43 & 69.96 & 68.44 & 67.07 & 66.09\\ 
{} & BM & 73.43 & 72.48 & 20.36 & 20.14 & 64.53 & 70.01 & \underline{80.52} & 79.40 & 70.19 & 67.35 & 69.01 & 67.63\\ 
{} & FBs & \underline{74.69} & 72.17 & \underline{23.82} & 22.79 & 71.49 & \underline{71.99} & 80.28 & 79.27 & 69.65 & 67.57 & 69.77 & 68.69\\ 
{} & FBc & 74.29 & 72.99 & 23.61 & 22.58 & \underline{71.86} & 71.61 & 80.37 & \underline{79.52} & \underline{70.70} & \underline{68.45} & 68.75 & 67.04\\ 
{} & RLB & 74.33 & \underline{73.26} & 23.77 & \underline{23.14} & 71.52 & 70.74 & 80.48 & \underline{79.52} & 70.13 & 68.11 & \underline{70.52} & \underline{70.48}\\\cline{2-14} 
{} & \methodname{} & \textbf{75.60} & \textbf{75.72} & \textbf{24.94} & \textbf{24.52} & \textbf{72.98} & \textbf{73.13} & \textbf{81.31} & \textbf{80.10} & \textbf{71.39} & \textbf{69.05} & \textbf{71.13} & \textbf{71.27}\\\hline

\multirow{9}*{400}  & Const & 71.06 & 68.34 & 17.09 & 16.96 & 64.01 & 58.93 & 78.49 & 77.98 & 68.19 & 66.69 & 68.66 & 68.33\\ 
{} & Rand & 70.05 & 67.74 & 20.90 & 20.45 & 64.25 & 60.58 & 78.62 & 78.43 & 68.88 & 67.64 & 70.36 & 69.75\\ 
{} & Bmub & 72.27 & 70.26 & 22.21 & 20.49 & 64.37 & 63.15 & 79.97 & 78.90 & 69.71 & 68.11 & 69.93 & 68.56\\ 
{} & Lin & 72.99 & 71.02 & 24.18 & 22.94 & 65.52 & 65.44 & 80.01 & 78.99 & 70.53 & 69.12 & 70.37 & 69.10\\ 
{} & BM & 74.96 & 73.01 & 25.59 & 23.74 & 65.87 & 68.38 & 80.90 & 79.91 & 71.62 & 70.35 & 71.58 & 70.44\\ 
{} & FBs & \underline{75.85} & 73.53 & 26.47 & 24.50 & 71.72 & 70.06 & 81.36 & 80.22 & 71.75 & 70.17 & 71.93 & \underline{70.85}\\ 
{} & FBc & 75.66 & 73.77 & 26.21 & 24.27 & 72.03 & 71.95 & 81.29 & 80.18 & 71.88 & 70.38 & 71.01 & 69.56\\ 
{} & RLB & 75.25 & \underline{74.96} & \underline{26.78} & \underline{24.83} & \underline{72.31} & \underline{72.24} & \underline{81.55} & \underline{80.47} & \underline{71.99} & \underline{70.59} & \underline{72.45} & 70.72\\\cline{2-14} 
{} & \methodname{} & \textbf{76.59} & \textbf{76.33} & \textbf{27.65} & \textbf{25.86} & \textbf{73.85} & \textbf{73.63} & \textbf{81.86} & \textbf{80.69} & \textbf{72.54} & \textbf{71.84} & \textbf{73.38} & \textbf{71.66}\\\hline

\multirow{9}*{600}  & Const & 71.05 & 69.36 & 23.10 & 21.66 & 64.61 & 61.77 & 79.28 & 78.49 & 68.39 & 67.01 & 69.21 & 68.69\\ 
{} & Rand & 68.79 & 69.05 & 22.72 & 20.32 & 64.39 & 62.49 & 79.25 & 78.83 & 69.31 & 67.95 & 70.19 & 69.74\\ 
{} & Bmub & 71.95 & 71.07 & 18.90 & 22.02 & 64.41 & 63.78 & 80.68 & 79.38 & 70.49 & 68.71 & 70.78 & 69.60\\ 
{} & Lin & 73.54 & 72.57 & 24.43 & 24.79 & 66.92 & 66.18 & 80.86 & 79.58 & 71.44 & 69.92 & 71.21 & 69.94\\ 
{} & BM & 75.25 & 73.58 & 28.30 & 26.62 & 67.21 & 67.80 & 81.42 & 80.26 & 72.47 & 71.07 & 71.97 & 70.82\\ 
{} & FBs & 76.18 & \underline{74.16} & 28.85 & 27.25 & 73.55 & 71.81 & 81.47 & 80.34 & 72.51 & 71.06 & 72.26 & 72.23\\ 
{} & FBc & \underline{76.25} & 73.98 & \underline{29.07} & 28.95 & \underline{74.14} & \underline{73.31} & 81.49 & 80.31 & 72.51 & 70.99 & 72.18 & \underline{72.84}\\ 
{} & RLB & 76.06 & 73.15 & 28.52 & \underline{29.60} & 73.85 & 73.05 & \underline{81.68} & \underline{80.60} & \underline{73.07} & \underline{71.64} & \underline{73.41} & 72.81\\\cline{2-14} 
{} & \methodname{} & \textbf{76.93} & \textbf{76.71} & \textbf{29.91} & \textbf{30.55} & \textbf{74.46} & \textbf{74.05} & \textbf{82.16} & \textbf{80.93} & \textbf{73.21} & \textbf{71.86} & \textbf{74.63} & \textbf{73.79}\\\hline

\multirow{9}*{800}  & Const & 67.21 & 66.43 & 23.63 & 21.95 & 68.17 & 64.97 & 79.64 & 78.81 & 68.85 & 67.49 & 69.49 & 69.01\\ 
{} & Rand & 68.95 & 71.02 & 24.54 & 20.66 & 68.15 & 65.32 & 79.78 & 79.23 & 70.13 & 68.75 & 70.91 & 70.11\\ 
{} & Bmub & 71.90 & 72.16 & 25.97 & 19.45 & 69.24 & 66.51 & 81.08 & 79.77 & 70.80 & 69.05 & 71.52 & 70.60\\ 
{} & Lin & 75.11 & 72.66 & 25.46 & 28.06 & 71.87 & 69.03 & 81.37 & 80.12 & 71.61 & 70.16 & 71.76 & 70.46\\ 
{} & BM & 75.28 & 73.89 & 28.76 & 29.00 & 72.83 & 70.31 & 81.64 & 80.58 & 72.89 & 71.63 & 73.09 & 71.74\\ 
{} & FBs & 76.09 & 75.04 & 29.54 & 30.18 & 75.92 & 73.86 & 81.87 & 80.83 & 72.99 & 71.72 & 73.42 & 72.20\\ 
{} & RLB & \underline{76.31} & \underline{76.34} & \underline{30.05} & \underline{30.81} & \underline{76.39} & \underline{74.72} & \underline{82.06} & \underline{81.07} & \underline{73.62} & \underline{72.37} & \underline{74.90} & \underline{73.18}\\\cline{2-14}
{} & \methodname{} & \textbf{77.29} & \textbf{76.78} & \textbf{32.82} & \textbf{32.46} & \textbf{77.10} & \textbf{75.57} & \textbf{82.47} & \textbf{82.69} & \textbf{73.77} & \textbf{73.55} & \textbf{75.39} & \textbf{73.82}\\\hline 
\end{tabular}
}
\label{tab:acc_non_iid}
\end{table*}

\section{Conclusions}
In this paper, we propose the Multi-session Budget Optimization Strategy for forward Auction-based Federated Learning (\methodname{}). \methodname{} is designed to empower FL model users with the ability to strategically allocate budgets over multiple FL training sessions and judiciously distribute the budget among data owners within each session by bidding with different bid prices, in order to maximize total utility. Based on the hierarchical federated learning, \methodname{} jointly optimizes inter-session budget pacing and intra-session bidding for model users in the auction-based federated learning ecosystem. Extensive experiments on six benchmark datasets have validated the effectiveness of \methodname{} in terms of utility gained and accuracy of the FL models.
To the best of our knowledge, it is the first budget optimization decision support method with budget pacing capability designed for MUs in multi-session forward auction-based federated learning.

\section*{Acknowledgments}
This research/project is supported by the National Research Foundation Singapore and DSO National Laboratories under the AI Singapore Programme (AISG Award No: AISG2-RP-2020-019); and the RIE 2020 Advanced Manufacturing and Engineering (AME) Programmatic Fund (No. A20G8b0102), Singapore.  

\bibliographystyle{IEEEtran}
\bibliography{main.bib}

\end{document}